\documentclass[acmsmall,manuscript,screen]{acmart}
\usepackage{times}
\usepackage{latexsym}

\usepackage[T1]{fontenc}

\usepackage[utf8]{inputenc}

\usepackage{microtype}

\usepackage{graphicx}
\usepackage{todonotes}
\usepackage{adjustbox}

\usepackage{enumitem}

\newcommand\example[1]{(\textcolor{blue}{\textit{#1}})}
\newcommand\newtext[1]{\textcolor{black}{#1}}

\newcommand\delidata{DeliData }

\AtBeginDocument{%
  \providecommand\BibTeX{{%
    \normalfont B\kern-0.5em{\scshape i\kern-0.25em b}\kern-0.8em\TeX}}}

\begin{document}

\title{DeliData: A dataset for deliberation in multi-party problem solving}

\author{Georgi Karadzhov}
\affiliation{%
  \institution{University of Cambridge}
  \city{Cambridge}
  \country{United Kingdom}}
\email{georgi.karadzhov@cl.cam.ac.uk}

\author{Tom Stafford}
\affiliation{%
  \institution{University of Sheffield}
  \city{Sheffield}
  \country{United Kingdom}}
\email{t.stafford@sheffield.ac.uk}

\author{Andreas Vlachos}
\affiliation{%
  \institution{University of Cambridge}
  \city{Cambridge}
  \country{United Kingdom}}
\email{av308@cam.ac.uk}
\begin{abstract}
 Group deliberation enables people to collaborate and solve problems, however, it is understudied due to a lack of resources.
To this end, we introduce the first publicly available dataset containing collaborative conversations on solving a \newtext{well-established} cognitive task, consisting of 500 group dialogues and 14k utterances. In 64\% of these conversations, the group members are able to find a better solution than they had identified individually, \newtext{and in 43.8\% of the groups who had a correct answer as their final solution, none of the participants had solved the task correctly by themselves}. Furthermore, we propose a novel annotation schema that captures deliberation cues and release all 14k utterances annotated with it. 
Finally, we use the proposed dataset to develop and evaluate two methods for generating deliberation utterances. 
The data collection platform, dataset and annotated corpus are publicly available at \url{https://delibot.xyz}.
\end{abstract}



\maketitle

\section{Introduction}


Group deliberation occurs in a variety of contexts, such as hiring panels, study groups, and scientific project meetings. 
It is traditionally explored in the field of psychology, where researchers examine the conditions under which a group can make better decisions. \citet{mercier2011humans} discuss how a group can outperform even the most knowledgeable individual within it -- \textit{the assembly bonus effect}.
This was also demonstrated by \citet{navajas2018aggregated} who showed that small focus groups can outperform the wisdom of the crowd.  

The aforementioned psychology research has mainly focused on the outcomes of the discussion, with less focus on analysing the discussion itself. 
However, the latter is necessary to understand what makes deliberation successful and inform interventions to facilitate it.
This is also echoed by \citet{vecchi-etal-2021-towards}, who identifies studying argumentation and deliberation as essential for the future of digital democracy.


In order to study what makes deliberations successful and learn how to intervene to this effect, we need a dataset that contains discussions where groups collaborate to solve a task. \newtext{Furthermore, the task should be such that the correctness of the decisions made}
can be objectively measured. 
Most existing datasets are between two interlocutors \citep{budzianowski2018multiwoz, dinan2019wizard, andersonmaptask}, thus not containing group discussions. Focusing on group datasets, one could consider negotiation dialogues \cite{afantenos2012modelling}, which, while multi-party, are adversarial in nature, therefore not containing collaboration. Publicly available datasets containing collaborative group discussions are WikiDisputes \citep{de2021beg} and AMI \cite{carletta2005ami}, but neither is associated with an objective measure of decision correctness, thus not enabling researchers to assess
how well did the conversation go.
\citet{niculae2016conversational} collected a group dataset containing collaborative problem-solving conversations with an objective \newtext{measurement of decision correctness}, but their dataset is not publicly available. 

\begin{figure*}[t!]
\centering
\includegraphics[scale=0.35]{"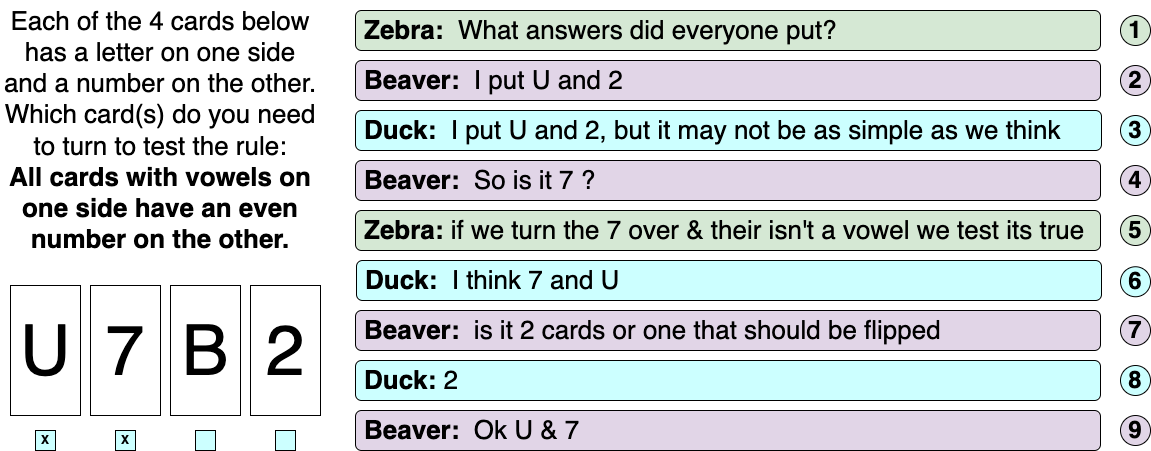"}
\caption{Abridged conversation from our dataset between 3 people solving the Wason card selection task}
\label{fig:delibot_interface}
\end{figure*}

In this work, we present the first publicly available dataset for group deliberation \newtext{associated with a quantitative measure of decision correctness}: 
\textbf{\delidata -- Deliberation Dataset}. An example conversation is in Figure~\ref{fig:delibot_interface}, with a group deliberating to solve the Wason card selection task \citep{wason1968reasoning}, a well-studied task in cognitive psychology. 
In the example, the group engages in various deliberation strategies: a participant is moderating the conversation by prompting the group for a response (utterance 1), whereas in utterance 4 a participant suggests exploring a different solution. Overall, the group starts with a common, but wrong, solution (utterances 2 and 3) and converges on the correct solution (utterances 6 and 9). 

We focus on the Wason card selection task as it is a well-studied task by researchers in the psychology of reasoning \cite{evans2016reasoning}. Furthermore, it does not rely on prior knowledge of participants and is well-characterised as demonstrating the benefits of group deliberation \cite{mercier2011humans}. This allows us to focus on the fundamentals of the deliberation process itself, and to study the factors which affect group \newtext{decision-making}: how individuals bring their knowledge and intuitions to the group, and how the exchange of arguments enables groups to combine information to navigate the problem space. \delidata allows us to 
test ideas on how this 
is done successfully, and thus suggest \newtext{appropriate} 
interventions. Using a task that requires no prior knowledge 
increases the chances that our findings are transferable to other domains of argument exchange,
such as education (e.g.\ tutorial groups), formal moderation (e.g.\ board meetings, policy discussions), and informal discussions (e.g.\ deciding where a group should get food in the evening).

The \delidata corpus contains 500 group dialogues, each of them \newtext{associated with a measurement of decision correctness (hereafter task performance)} before and after the group discussion. Given these measurements, we show that after discussing the solution, 64\% of the groups perform better at the Wason task, compared to the performance of their members individually. Moreover, in 43.8\% of the groups who had a correct answer as their final solution, none of the participants had solved the task correctly by themselves, thus demonstrating how people can solve the task better through deliberation. 
To aid future analysis and dialogue system development we propose an annotation schema that captures conversational dynamics and deliberation cues in collaborative conversations, and release the 500 dialogues as an annotated corpus with 14k annotated conversational turns in total.
%
Further, we showcase the multiple possible uses of the DeliData corpus by conducting a wide range of analyses and modelling experiments, including predicting whether the deliberation improves the decision-making and the generation of utterances that can probe the conversation by asking questions.
Finally, we demonstrate the generalisability of the annotation schema and the annotated dataset by automatically annotating a real-world collaborative task - a group of people debunking deep fake images.

\section{Related Work}


\citet{niculae2016conversational} investigated group collaboration in the context of playing a game attempting to geo-locate a photo on the map. In their experimental setup, they first evaluate each participant individually, and after that, they initiate a group discussion and finally ask the group to make a decision together. Unfortunately, their dataset is not publicly available, and thus cannot be used in other studies.
Likewise, \citet{debatebot} investigates how groups of people collaborate in solving a task together, as well as how a dialogue system can be incorporated within the discussion. Unfortunately, their dataset contains only 12 discussions, making it too small for any reasonable analysis or dialogue system development, and similarly to the dataset of \citet{niculae2016conversational}, their dataset is also not publicly available. 

Wikipedia is a popular source of collaborative conversations. 
\citet{hua2018wikiconv} collect 91M discussions from Wikipedia, together with the  edits discussed in them. It is the largest dataset that captures group collaboration, but it is not supported by an annotated corpus.
This is partly addressed by \citet{al-khatib-etal-2018-modeling}, who annotate 
200k discussion turns from Wikipedia in 33 dimensions based on discourse acts, argumentative relations and semantic frames. 
However, unlike the conversations of
\citet{niculae2016conversational} and the work presented in this paper, there is no assessment of whether the participants in a conversation on Wikipedia reached a better decision,
which renders assessing constructiveness more difficult because there is usually no objectively correct answer.

\par

Related to constructive conversations is the research on negotiation dialogues which have been explored in the context of games \citep{keizer2017evaluating, cuayahuitl2015strategic} and trading \citep{hehe-negotiation, deal-nodeal}. 
However, even though negotiation dialogue research often deals with multiparty conversations \cite{cuayahuitl2015strategic}, such systems are by nature adversarial, rather than constructive. 

Multiparty conversations are also the focus of \citet{carletta2005ami}, who created a multi-modal corpus of business meetings containing audio, video, transcriptions and auxiliary materials provided to the participants. However, they did not explore deliberation strategies, nor tried to measure the productivity of the group. 
Using parts of this dataset, 
the CALO project \cite{tur2010calo} proposed a toolkit to assist group meetings, such as dialogue act segmentation, action item recognition and others, but no attempt to assess constructiveness was made. Similarly, previous research \citep{zeng2020changed, hidey-etal-2017-analyzing} has investigated how people change their minds in online forums. But as the topics in online forums are very complex and the data is noisy, there is no objective measure of constructiveness.
Finally, \citet{de2019learning} evaluated turn prediction in the context of group dialogues. They evaluate their system on 3 datasets: one is proprietary, one is artificially created by combining 1-to-1 dialogues from \citet{budzianowski2018multiwoz},
and the third dataset consists of transcripts of a popular TV show, which while containing true multi-party dialogues, are not collaborative.

\section{Experimental Setup}
\label{sec:experimental_setup}

In our experiments with the Wason card selection task \cite{wason1968reasoning}, participants are presented with 4 cards with a number or a letter on them. They have to answer the following question ``\textit{Which cards should you turn to test the rule:} \textbf{All cards with vowels on one side have an even number on the other?}''. 
\newtext{A common fallacy} is to 
select the vowel and the even number (i.e.\  selecting the two cards mentioned in the question), which is incorrect, demonstrating \textit{confirmation bias}  \citep{mercier2011humans}. The correct answer is to turn the vowel, to check for an even number on the other side, and to turn the odd number, to verify there isn't a vowel on the other side. 
\par
We calculate \textbf{task performance} in two ways. First, we consider a \textbf{coarse-grained} (binary) scoring of the task - \textbf{Correct - 1} if the vowel and odd number are selected, \textbf{Incorrect - 0} otherwise. Recognising that the coarse-grained scoring may needlessly penalise answers that are close to the correct one, we also devised an alternative \textbf{fine-grained} scoring. We grant 0.25 points for (i) turning a vowel or an odd number, and (ii) for \textbf{not} turning the even number or the consonant. Therefore, if the participant submitted a correct solution, their score would be 1, if they are off by one card - 0.75 and so on. We also calculate \textbf{performance gain}, by subtracting the average of the solo solutions from the average of the group performance. For example, if the average score of participants' solo submissions was 0.5 and 
improved to 0.75 after the discussion, the group performance gain would be $0.75 - 0.5 = 0.25$. 
We collect the data using the following protocol (full participant instructions available \newtext{in} Appendix \ref{app:reproducibility_insructions}):
\begin{enumerate}[noitemsep,nolistsep]
\item \textbf{Solo Phase}. Each of the participants in the group is presented with the same 4 cards and submits a solution to the task.
\item \textbf{Group Phase}. Following the solo phase solution submission,
participants gain access to a chatbox to 
share their solutions and discuss. We encourage them to do so
for at least 5 minutes
but no longer than 7 minutes without enforcing these time limits; thus there are cases with very short and very long conversations. 
\item \textbf{Revised Submission}. After discussing their solutions, the participants are asked to revise their initial card selection and submit again. 
\end{enumerate}

We posted our data collection on the crowd-sourcing platform Mechanical Turk 
with the following job specification:
\begin{enumerate}[noitemsep,nolistsep]
    \item Everyone who completes the task is paid \$2.00 (approx. £1.60). Participants are given a bonus of \$1.00 (£0.80) if they 
    return the right answer.
    As the average time for participation is about 8 minutes, each participant is paid £12/hour (or £18/hour if they solve the task correctly). This is between 35\% and 102\% above UK's National Living Wage \footnote{£8.91/hour as of 01/04/2021, based on \url{https://www.gov.uk/government/publications/the-national-minimum-wage-in-2021}}.  
    \item No personal information is collected and the participants are asked not to share anything that may reveal personal details.
    \item We recruited only adult participants from countries where English is a primary language, and they complete a simple reading comprehension test. The only language used in our dataset is English.   
\end{enumerate}
 Participants are informed that we are investigating how people collaborate in solving a cognitive task and that we will be saving chat transcripts. This experimental protocol was approved by the ethics committee of the authors' institution.
 
\par
The data collection is performed using a web application we call \textit{DialogueDen}, which we open-source together with this study. 
The design of the platform allows us to record solo and group selections and the state of the game \newtext{at} key points of the experiment. This data can be used to identify when a participant reached the correct decision, even if they don't express it explicitly in the chat. 
Moreover, we integrated a number of features to \textit{DialogueDen} that are specific for the data collection on Mechanical Turk, addressing various issues that arise when collecting group conversations in an unsupervised manner. These are part of the code release and are presented in detail in Appendix \ref{app:mturk}. 
\begin{table}[t!]
    \centering
    \begin{tabular}{|l|c|c|c|}
    \hline 
    \textbf{} & \textbf{Pilot} & \textbf{Mturk} & \textbf{Total} \\
\hline 
    Number of Dialogues & 18 & 482 & 500 \\ \hline 
    Total Participants & 53 & 1526 & 1579 \\ \hline
Total number of utterances & 705 & 13298 & 14003 \\ \hline
 AVG  utterances 
    
    & 39.2 & 27.6 &  28  \\ \hline
 
    AVG utterance length & 8.19 & 8.62 & 8.59 \\ \hline
AVG unique tokens 
    
    & 78.1 & 67.6  & 68  \\ \hline 
AVG number of participants 
     & 2.94 & 3.17 & 3.16 \\ \hline 
     Solo Performance (fine-grained) & 0.59 & 0.59 & 0.59 \\ \hline 
     Group performance (fine-grained) & 0.81 & 0.71 & 0.72 \\ \hline 
     Solo Performance (coarse-grained) & 0.19 & 0.11 & 0.11  \\ \hline
    Group performance (coarse-grained) & 0.57 & 0.32 & 0.33 \\ \hline 
    AVG group agreement & 0.92 & 0.83 & 0.83 \\ \hline 
\end{tabular}
\caption{Corpus statistics for pilot and MTurk data.}
\label{table:initial_data_collection}
\end{table}

\section{\delidata dataset}
\label{sec:delidata}

Using the experimental protocol above we initially conducted a pilot study, where we collected 18 group dialogues, with 53 volunteers from a university psychology department, who didn't have prior knowledge of the task.
After that, we ran a larger scale data collection on Mechanical Turk which is often used for data collection in behavioural research and often produces similar results to in-lab experiments  \cite{crump2013evaluating}. 
This data collection was not moderated in any way, making it a realistic data collection process. We ensure the quality and anonymity of the data from MTurk by manually checking each conversation. We excluded a total of 160 conversations that were too short, of poor quality or with too few actively engaged participants. Thus, we release 482 dialogues that are of comparable quality to our in-lab pilot.

Summarised statistics of the two subsets are presented in Table \ref{table:initial_data_collection}. While the two subsets differ in terms of absolute performance, the improvement from solo to group performance is substantial in both data collections for both coarse- and fine-grained metrics,  
in agreement with results from psychology research on offline deliberation \citep{mercier2011humans}, and thus validating our data collection approach using MTurk.
Another difference is that the average number of utterances per dialogue is lower on MTurk, which we attribute to the psychology student volunteers being more dedicated than crowd workers.

\par

In Table~\ref{table:corpora_comparison} we compare three multi-party dialogue datasets: StreetCrowd \cite{niculae2016conversational}, Settlers of Catan (SoC) \cite{afantenos2012modelling}, and ours. Of these three, only two are collaborative - ours and StreetCrowd, as SoC is among players competing against each other. 
Ours is the only one containing collaborative group conversations available for research. Moreover, while it contains fewer dialogues than StreetCrowd, these are 2.5 times longer in terms of utterances, thus more likely to exhibit collaborative strategies spanning over multiple utterances.



\begin{table}[t!]
    \centering
    \begin{tabular}{|c|c|c|c|}
    \hline
    \textbf{Property} 
    & \textbf{StreetCrowd} 
    & \textbf{SoC} 
    & \textbf{\delidata} 
    \\ \hline
       dialogues  & 1,450 & 32 & 500  \\ \hline
       utterances  & 17,545 & 2,512 & 14,003  \\ \hline
    \multicolumn{1}{|p{2cm}|}{\centering utterances per dialogue}  & 12.1 & 78.5 & 28  \\ \hline
    \multicolumn{1}{|p{2cm}|}{\centering utterance length} & 5.33 & N/A & 8.59  \\ \hline
    \multicolumn{1}{|p{2cm}|}{\centering pub. available} & No & No & Yes  \\ \hline
    collaborative & Yes & No & Yes  \\ \hline
    annotated & No & No & Yes  \\ \hline
    \end{tabular}
    \caption{Multiparty dialogue corpora comparison}
    \label{table:corpora_comparison}
\end{table}

\section{Annotating deliberation cues}
\label{sec:annotation}

\begin{figure*}[htb!]
\centering
\includegraphics[scale=0.4]{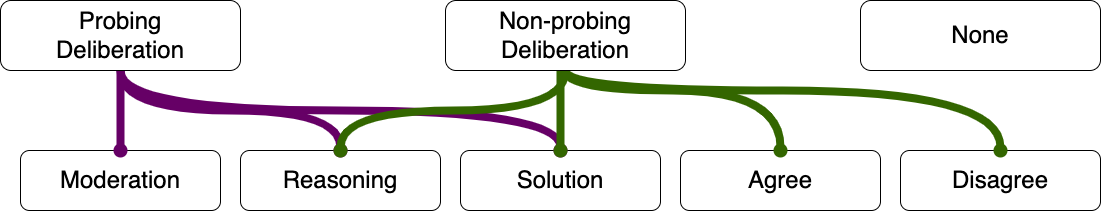}
\caption{Hierarchical annotation structure}
\label{fig:annotation_structure}
\end{figure*}

The DeliData corpus introduced in the previous sections contains multi-party discussions of people solving the Wason card selection task. The transcripts are augmented with some metadata, such as when someone clicked on a card or submitted a new solution via the interface. Both the transcripts and the metadata can be used for various tasks, such as dialogue systems training, evaluation of conversation performance and solution-finding patterns. That said, in order to enable more sophisticated analysis and application in the study of deliberation, a fine-grained annotation of each conversation turn would be required. In this section, we introduce DeliAnnotation - an annotation scheme designed to study deliberation in collaborative problem-solving.

\subsection{Desiderata and previous work}

In order to annotate DeliData, we draw inspiration from theoretical work on argumentation \citep{walton_2005, toulmin_2003}, and from studies investigating how people reason and deliberate together to achieve a common goal \cite{mercier2011humans}.

Given these previous studies, as well as ensuring the transferability of the annotated data we outline 3 key criteria that an annotation scheme should fulfil: 
\begin{itemize}
    \item The annotation scheme should capture general argumentation structure, e.g.\ distinguishing between arguments about reason and arguments about a solution. Simple interactions like agreement and disagreement should also be captured.
    \item The annotations should highlight deliberative cues such as moderation, argument probing, and solution management. 
    \item The annotation scheme should be specific enough to capture deliberation and collaborative problem-solving phenomena, but should be general enough to be applied to a range of problem-solving tasks.
\end{itemize}

Given these desiderata, we first considered using the annotation schema previously proposed for discourse parsing \cite{coarse-discourse} and Wikipedia discussions \cite{al-khatib-etal-2018-modeling}. Both of these schemata capture some discussion markers, such as questions, argumentation and agreement, which are important in analysing constructive discussions. Unfortunately, neither of them 
capture
how people collaborate, which could be achieved by identifying deliberative cues. Furthermore, these schemata carry over some specific labels related to discourse parsing and Wikipedia editing. 
In terms of collaborative discussions, the MapTask schema by \citet{carletta1996hcrc} annotates conversations between two participants, who play a game together. Their annotation scheme is limited to basic interactions such as questions and answers,  
and it is
missing important deliberative queues such as probing, argumentation and solution tracking.

\subsection{DeliAnnotation}

We propose an annotation schema with 3 levels of annotation, each focusing on different aspects of deliberation.
Figure~\ref{fig:annotation_structure} gives the overview of the schema, and we describe it in detail in the remainder of this section.

At the top level of the schema, we are interested in identifying the \textbf{probing deliberation}, i.e.\ any utterance that provokes discussion, deliberation or argumentation \textit{without} introducing novel information \example{Hey, @Cat what do you think was the solution?}. 
Such utterances could be considered conversational interventions that may change the flow of the conversation to induce further arguments or to moderate a conversation. On Figure~\ref{fig:delibot_interface} these would be utterances 1, 4, and 7.
We also recognise that most utterances in a conversation are not probing, but are inherently useful for the conversation. We label these utterances as \textbf{non-probing deliberation} (also abbreviated as NPD), and they include all discussions that are concerned with the task's solution and participants' reasoning \example{I think the answer is A, because we have to check each vowel for sure}. 
Finally, we include a \textbf{None} label that covers all utterances that are not related to the previous two categories. These utterances often include familiarities \example{Greetings fellas} or hesitation cues \example{hmm...}.
We refer to the first level of annotation as \textbf{Type}. After distinguishing between probing and non-probing deliberation, we classify each utterance into 5 roles at the second level:
\begin{itemize}[noitemsep,nolistsep]
\item \textbf{Moderation} (exclusive to probing deliberation): Moderation utterances are not concerned directly with the task at hand, but rather with \textit{how} participants converse about it \example{Let's discuss our initial solutions}. These utterances 
are concerned with \newtext{modelling} the conversation dynamics.
\item \textbf{Reasoning}: Utterances focusing on argumentation and can be both probing \example{Why did you think it wasn't 8?} and non-probing  \example{I think it would be 7 to test if it would be incorrect}. 
\item \textbf{Solution}: Utterances that are managing the solution of the task. Can be both probing \example{Are we going for A and 4?} or non-probing  \example{I think the answer is 7 and A}. 
\item \textbf{Agree} and \textbf{Disagree} (exclusive to non-probing-deliberation): Utterances expressing agreement or disagreement with a previous argument or solution. 
\end{itemize}
An important caveat with \textbf{Reasoning} is that it takes a priority over other labels. 

Some of the utterances may carry additional information beyond what is captured by their type and role, i.e.\ the first two levels of the annotation.
Therefore, we introduce a set of \textbf{additional labels} 
that mark specific phenomena in the conversation, which we defined as follows: 
\begin{itemize}[noitemsep,nolistsep]
    \item \textbf{specific\_addressee}: Utterances explicitly addressing specific participant(s) \example{@Llama what do you think?}
    \item \textbf{complete\_solution} and \textbf{partial\_solution}: Utterances 
    advocating for either a complete task solution \example{Let's turn A and 7}, or a partial one \example{one of the cards is A}. 
    
    \item \textbf{solution\_summary}: Utterances that recall previous solutions to prompt for an agreement
    \example{So, do we all agree on A and 5?}.
    
    \item \textbf{consider\_opposite} - utterance suggesting an opposite solution. \example{maybe not L?}
\end{itemize}

\subsection{Annotated dataset}
Using the annotation schema introduced in this section we annotated all dialogues presented in section~\ref{sec:delidata}.
We performed an annotation agreement study between 3 annotators on 41 of the dialogues using Cohen's kappa \cite{cohen1960coefficient}. We obtained an inter-annotator agreement of 0.75 on the first level, 0.71 on the second level, and an average agreement of 0.53 on the additional labels.
\par
The label distribution for the first two levels is presented in Table~ \ref{table:annotated_distribution}. 
Overall, the number of \textbf{Reasoning} and \textbf{Solution} utterances are substantial,
confirming that the subjects in our data collection
engaged in substantial discussions about the solutions and their reasoning.
The corpus also contains 1739 \textbf{Probing} utterances, where most of which 
are \textbf{Moderation}. On the other hand, probing for Reasoning or Solution is fairly evenly distributed. This suggests that the strategies chosen for annotation are commonly used.
Finally, 3267 utterances were annotated as
non-deliberative (\textbf{``None''}).

In Table~\ref{table:additional_distribution} we present the distribution of additional labels. In column \textbf{Count} we show the total number of occurrences of each of these labels, while in \textbf{Prevalence} we show how often this label occurs in \textit{all} utterances, including those without annotation for an additional label. The most prevalent label is \textbf{complete\_solution}, appearing in about 20\% of the utterances. While the other additional labels occur less in the conversation (around 5\% or less), they might be useful for dialogue analysis.

\begin{table}[t!]

\centering
    \begin{tabular}{|l||l|l|l||l|}
    \hline 
    \textbf{} & \textbf{Probing} & \multicolumn{1}{|p{2cm}|}{\centering \textbf{Non-probing deliberation}} & \textbf{None} & \textbf{Total} \\
\hline \hline 
    Moderation & 785 & 0 & & 785 \\ \hline 
    Reasoning & 464 & 3920 & & 4384 \\ \hline 
    Solution & 490 & 2918 & & 3408 \\ \hline 
    Agree & 0 & 2048 & & 2048 \\ \hline 
    Disagree & 0 & 111 & & 111 \\ \hline \hline 
    \textbf{Total} & 1739 & 8997 & 3267 & 14003 \\ \hline
\end{tabular}
\caption{Frequencies for the labels in the top two levels of the annotation schema}
\label{table:annotated_distribution}
\end{table}

\begin{table}[t!]
\centering
    \begin{tabular}{|l|l|l|}
    \hline 
    \textbf{Additional Label} & \textbf{Count} & \textbf{Prevalence} \\
\hline \hline 
    specific\_addressee & 397 & 2.8 \% \\ \hline 
    complete\_solution & 2841 & 20.3 \% \\ \hline 
    partial\_solution  & 460 & 3.3 \% \\ \hline 
    solution\_summary & 367 & 2.6 \% \\ \hline 
    consider\_opposite & 45 & 0.3 \% \\ \hline 
\end{tabular}
\caption{Label distribution the additional labels}
\label{table:additional_distribution}
\end{table}

\section{Analysis}
\label{sec:analysis}
\subsection{Two-party and multi-party conversations}

\begin{figure}[!t]
\centering
\includegraphics[scale=0.5]{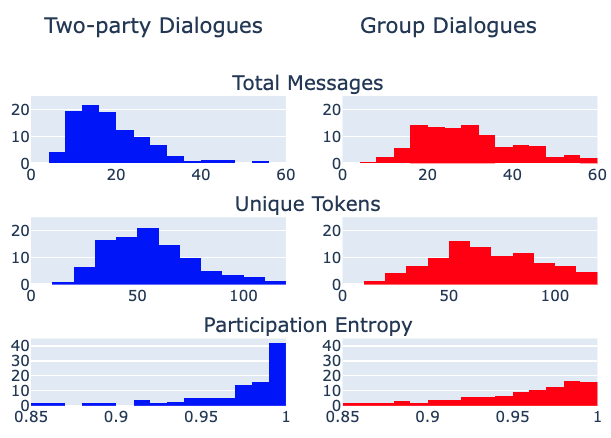}
\caption{Comparison between conversational statistics of two-party dialogues(left) and group dialogues (right). Each of the histograms is showing the percentage of dialogues on the y-axis.}
\label{fig:dyadsvtriads}
\end{figure}

While in our dataset two-party and multi-party (3 or more participants) conversations have similar statistics, there are notable differences that we highlight in this section. 
In Figure~\ref{fig:dyadsvtriads}, we present histograms comparing three conversational statistics - the total number of messages, number of unique tokens and participation balance, represented by entropy. First, dialogues between two interlocutors have mostly between 10 and 25 utterances, while group discussions in DeliData are 
\newtext{distributed}
in a wider range, between 20 and 40 utterances, with a long tail of conversations longer than 50 utterances. This  occurs naturally, as multiparty discussions contain more arguments and exchange of ideas. Likewise, participants in these discussions tend to use a larger vocabulary of words, 
as shown on the histograms of the unique tokens. 

In this analysis, we also look at how balanced the conversations are, i.e.\ whether all of the participants contributed equally. We calculate the participation entropy similarly to \citet{niculae2016conversational}, where 
\newtext{it}
is maximised if everyone participates equally, and approaches 0 if there is a large imbalance. In our dataset, the balance for two-party conversations is better, where 40 \% of the discussions are almost uniformly balanced, while in multi-party discussions, it is often the case that one of the participants is driving the discussion. This is not surprising, as in one-to-one conversations if one of the participants asks a question, it is customary that the other participant answers. Such is not the case for multi-party discussions, where some of the participants may decide to have a more passive role.

\par

Besides conversation statistics, we analyse 
the difference in task performance. Verifying for the initial conditions first, the solo performance of both types of groups is comparable - 0.597 and 0.585. On the other hand, the collective performance of these groups was 0.694 for two-party conversations and 0.724 for multi-party, thus the performance gain is 0.096 and 0.139 respectively. Therefore, we argue that it is the multi-party (as opposed to two-party) discussion that led to \newtext{an} improved conversational performance. \newtext{A limitation of this analysis is that we did not study whether increasing the number of group participants stops increasing performance gain. We anticipate that given previous findings by \citet{navajas2018aggregated} that communication among group members is important, the benefits from increasing the number of participants will be diminishing as communication among them and exchange of ideas becomes harder and the collaboration resembles the wisdom of crowds.}

\subsection{Propagation of correct solutions}

Analysing our data we found out that there is 0.36 Kendall's Tau B correlation \citep{kendall} between group consensus and performance gain.
%
An investigation of how correct solutions propagate through the conversations showed that 21.2\% of conversations started and finished with the same amount of correct submissions, thus the participants didn't convince anyone of the correctness of their responses. In 35\% of the discussions where a single participant had answered correctly in their solo submission, they convinced at least one more participant in the group phase. However the reverse also happened - in 4\% of all dialogues, the group convinced a participant with the correct answer to change it, which is considerably rarer than changing to the correct solution. 
Finally, in 43.8\% of the groups in which at least one participant submitted a correct response after the conversation, no participant had submitted a correct solution in their solo phase. This supports
the \textit{group is better than the sum of its parts} hypothesis, suggesting that deliberation offers more than just facilitating the spread of a correct solution among group members, and is consistent with the findings of \citet{moshman1998collaborative}  and \citet{schulz2006group}, who show that deliberation plays a bigger role in task success, compared to individual participants' ability. 

Furthermore, we present an analysis of different solution propagation patterns based on the annotation schema. We compared the groups where at least one of the participants had the correct solution in their solo phase, to the groups which reach the correct solution without anyone knowing the solution in their solo phase (referred to as DELI). The DELI subset contains a higher percentage of probing (17.3\% vs 14.4\%), and reasoning (43.8\% vs 37.8\%) utterances, suggesting that the participants are actively engaging in deliberation 
to get to the correct solution. Naturally, the DELI subset contains fewer utterances that propose a solution (30.4\% vs 35.7\%), as participants are more engaged with the reasoning behind the solution, \newtext{as} opposed to the solution itself. These findings 
are suggestive of the rich source of information about the dynamics of deliberation 
present in the data.

\section{Modeling Experiments}
\label{sec:modeling}
\subsection{Predicting conversation's performance gain}

\begin{table}[th!]
\centering
    \begin{tabular}{|l|l|l|}
    \hline 
    \textbf{Features} & \textbf{AUC} & \textbf{Stability} \\
\hline 
    (0) Majority Baseline & 0.5 &  \\ \hline 
    (1) Annotation Statistics & 0.52 & 1 \\ \hline
    (2) Annotation Role n-grams & 0.53 & 0.94 \\ \hline
    (3) SC Interaction  & 0.49 & 0.848 \\ \hline 
    (4) SC Linguistic  & 0.57 & 0.975\\ \hline 
    (5) Participation Dynamics & 0.61 & 0.886 \\ \hline
    (6) Conversational Statistics & 0.65 & 0.997 \\ \hline 
    (7) Annotation Type n-grams & 0.65 & 0.98 \\ \hline 
    (2) + (3) + (5) + (6) & 0.70 & 1 \\ \hline 
\end{tabular}
\caption{Predicting conversational performance gain}
\label{table:predicting_outcome_results}
\end{table}

In order to analyse the factors that make a conversation constructive, as well as showcase possible applications of the \delidata corpus, we perform a series of modelling experiments, where we predict the \textit{performance gain} of a conversation, i.e.\ whether group task performance improved or not following the deliberation.

In these experiments we use a simple decision tree classifier \cite{scikit-learn} with a maximum depth of 7 and minimum samples per leaf set to 5 and use leave-one-out cross-validation (LOOCV).
As the dataset is imbalanced (318 conversations with performance gain and 182 without),
we evaluate our models using the area under the ROC curve and stability. For these experiments, we considered 7 types of features: (1) Annotation statistics (i.e. normalised counts of each of the annotation labels), (2) n-grams of annotation Role sequences, (3) interaction features borrowed from StreetCrowd~\cite{niculae2016conversational} (SC Interaction) (4) linguistic features borrowed from StreetCrowd~\cite{niculae2016conversational} (SC Linguistic), (5) participation dynamics (i.e.\ whether one of the participants dominated the conversation),  (6) conversational statistics (number of messages, tokens, etc.), and (7) n-grams of annotation Type sequences. Full experimental details can be found in Appendix~\ref{app:reproducibility_performance_gain} and the code will be made publicly available.
As shown in Table~\ref{table:predicting_outcome_results}, the SC interaction features don't perform well in our setup if used alone, achieving accuracy that is below the baseline. Following, without feature combinations, both conversational statistics and Annotation Type n-grams are a good predictor of conversational performance. In terms of best overall performance, the best feature combination is achieved by using the interaction features from StreetCrowd, the participation dynamics, the conversational statistics, and the Annotation Role n-grams. Both SC Interaction and Participation Dynamics, model how participants interact with each other, providing a glimpse into group collaboration. On the other hand, we show that information encoded by deliberation annotation is also important for predicting performance gain. These results suggest that conversational dynamics are a strong addition to traditional feature-based approaches for dialogue classification.

\subsection{Modeling the annotation scheme}
\label{sec:delidata_prediction}

Given that annotating datasets is an expensive, time-consuming, and non-trivial endeavour, we propose to learn models that classify utterances using the annotation scheme proposed in the previous sections. 
Given that our annotation scheme is hierarchical, we considered \newtext{two} options - (i) build a single classifier that predicts both layers jointly, or (ii) build \newtext{three} classifiers that predict each level. For the latter case, we first classify whether an utterance is Probing, Non-probing deliberation (NPD) or None. Following, depending on the result of the first level classification, we have \newtext{one} model that classifies Probing roles, and one that predicts Non-probing roles. 

%

In terms of models, we evaluate two approaches for current utterance prediction. First, as a baseline method, we consider a TF-IDF encoding of the utterance, which is then passed to a random forest classifier with 5 estimators \cite{scikit-learn}.
Second, we experiment with a neural model, that relies on a pre-trained encoder for the utterance embedding. Each utterance is encoded by the GTR model \cite{ni2021large}, a T5-based encoder. Then each utterance encoding is passed to a fully-connected feed-forward neural network with one hidden layer of size 512, and one output layer with the desired number of outputs (depending on the level of annotation). The neural network is trained using an Adam optimizer \cite{kingma2015adam}.

\begin{table*}[h!]
\begin{tabular}{|l|l|l|l|l|}
\hline
\textbf{Model} & \textbf{Type} & \textbf{Role-Probing} & \textbf{Role-Non-probing} & \textbf{Combined Performance} \\ \hline
Majority Baseline & 0.64 & 0.45 & 0.44 & 0.25 \\ \hline
RandomForest & 0.76 & 0.59 & 0.73 & 0.40 \\ \hline
RandomForest-SC & - & - & - & 0.57 \\ \hline  
Neural & \textbf{0.87} & \textbf{0.86} &
\textbf{0.86} & \textbf{0.80} \\ \hline
Neural-SC & - & - & - & 0.79 \\ \hline  

\end{tabular}
    \caption{Classification performance of predicting the current utterance. Accuracy score reported. SC stands for a single classifier, where a single model is used to predict all labels. In all other cases, a cascading prediction of 3 models is used.}
    \label{tab:current_utterance_prediction}
\end{table*}

The comparison between the two classifiers is presented on table~\ref{tab:current_utterance_prediction}. For both methods, we kept the same experimental setup - 10-fold cross-validation, comparing accuracy. Both the RandomForest and the Neural model outperform the majority class baseline. 
In the column Combined Performance we compare the cascading classifier and the single (joint) classifier.
We can see that the RandomForest classifiers perform better in a single classifier setup while for the neural model, the combined performances are comparable. The three-classifier cascade version is slightly better, achieving \newtext{an} accuracy of 0.87 on the first level of annotation. The performance on the second level of annotation is 0.86 irrespective of whether the type of utterance is probing or not.


\subsection{Generating Probing Utterances}
\label{ssec:generation}


We conclude by developing and evaluating two methods for generating probing utterances. 
We consider
two different approaches - a retrieval-based approach and a generative approach with language models.
The task setup is: given the previous dialogue utterances and the \textbf{Role} of a probing utterance (i.e.\ Probing-Moderation, Probing-Reasoning, Probing-Solution), generate the most appropriate utterance to continue the dialogue.
For these experiments, we consider 50 of the annotated dialogues using the annotation schema of Section~\ref{sec:annotation} as we assume the Role of the utterance to be generated given,
and split them into a training set of 30 dialogues and a test set of 20.
In our experiments, we compare 4 candidate responses:

\begin{itemize}[noitemsep,nolistsep]
    \item \textbf{Original.} We take the utterance by the human participant from the original dataset.
    \item \textbf{Random.} We sample from the training data a random utterance that has the same \textbf{Role} as the one we need to generate. This is a strong baseline, as sampling for the same role often yields a contextually adequate utterance (albeit not necessarily the best). 
    \item \textbf{Retrieval.} We find the most similar utterance with the same Role in our training dataset. To calculate similarity we encode the context of the probing utterance using a \newtext{pre-trained} DialoGPT model  
    \item \textbf{Generative} We use a \newtext{pre-trained} DialoGPT to generate the next utterance based on the current conversation context.
\end{itemize}

\begin{table*}[t!]
    \centering
    \begin{tabular}{|c|l|}
        \hline
        \textbf{Context} & 
        \multicolumn{1}{p{10cm}|}{but if we are trying to verify then maybe we select them all} \\ \hline
         \textbf{Original} & how else could you know? \\ \hline
         \textbf{Random} & Why did you press V \\ \hline
         \textbf{Retrieval} & \multicolumn{1}{p{10cm}|}{How many cards do you think at minimum we need to flip to confirm the rule} \\ \hline
         \textbf{Generative} & \multicolumn{1}{p{10cm}|}{I think he means that the list of possible candidates is a list that will be evaluated in the upcoming days.} \\ \hline
    \end{tabular}
    \caption{Utterances generated by different methods}
    \label{tab:generative_example}
\end{table*}

For every method (except for the original) we replaced with placeholders both the mentions of participants and solutions.
Once we generate an utterance, if it has a mention of a participant or a solution, we use a simple rule-based system to select \newtext{the} appropriate substitution from the context. We show an abridged example from our  experiments in Table~\ref{tab:generative_example} (additional examples in Appendix~\ref{app:example_NLG}). 
We evaluate the three generated candidate responses using both automatic and human evaluation.

\begin{table}[t!]
\centering
    \begin{tabular}{|l|l|l|l|}
    \hline 
    \textbf{Method} & \textbf{BLEU-4} & \textbf{Similarity} &
    \multicolumn{1}{|p{1cm}|}{\centering \textbf{BERT Score}}\\
\hline 
    Retrieval & 0.39 & 0.56 & 0.83  \\ \hline 
    Random & 0.35 & 0.55 & 0.83  \\ \hline 
    Generative & 0.09 & 0.42 & 0.79  \\ \hline 
\end{tabular}
\caption{Automatic evaluation of Probing generation }
\label{table:nlg_auto}
\end{table}

\begin{table}[t!]
    \centering
    \resizebox{25em}{!}{
    \begin{tabular}{|c|c|c|c|c|}
    \hline
         \textbf{Original} & \textbf{Retrieval} & \textbf{Random} & \textbf{Generative} &  \\ \hline
          - & 0.5 & 0.46 & 0.28 & \textbf{Original} \\ \hline
         0.5 & - & 0.48 & 0.29 & \textbf{Retrieval} \\ \hline
          0.54 & 0.52 & - & 0.27 & \textbf{Random} \\ \hline
          0.72 & 0.71 & 0.73 & - & \textbf{Generative} \\ \hline
    \end{tabular}
    }
    \caption{The table reports pairwise preferences in columns over rows, i.e.\ the first column reports the preference of the Original text vs the other 3 methods.}
    \label{tab:preference}
\end{table}

First, we applied three commonly used measures for evaluating NLG applications - BLEU 4 \cite{bleu}, sentence similarity using RoBERTa \cite{Liu2019RoBERTaAR}, and BERTScore \cite{bertscore}. As none of our NLG methods is trained to generate the \textit{same} utterance as the Original, we do not expect that any of the candidate responses will achieve strong results, but automatic measures for NLG evaluation can be a good proxy for the quality of generated responses.
On Table~\ref{table:nlg_auto}, we present the results where we compare them to the Original response. The \textbf{Retrieval} approach has the best overall performance, with a BLEU-4 score of 0.39 compared to 0.35 and 0.09. If we consider just the Similarity and BertScore measures, the \textbf{Retrieval} and \textbf{Random} approaches have similar performance. On the other hand, 
\textbf{Generative} 
performs consistently worse on all measures. 

We also perform a human evaluation study, where we asked people to rate the generated responses. We recruited 28 workers from Prolific using comparable worker qualifications and payment levels as on MechanicalTurk. We gave crowd workers the following instructions: ``Please rank the 4 candidate responses from 1 (for the best response) to 4 (for the worst). You can give the same rank for responses you consider equally good/bad by placing them in the same box.''. We asked each of the crowd workers to rank 10 sets of candidate responses, which resulted in 280 annotations of 89 probing cases. First, we compared the average ranks of each of the NLG methods. The Original and the Retrieval approaches had similar ranks - 2.12 and 2.15, while the Random candidate was ranked on average at 2.23. Finally, the generative approach performed the worst, being ranked on average at 3.02. 
To gain a more fine-grained understanding of which method is preferable, we 
calculated the pairwise preferences (adjusted for ties), presented in Table~\ref{tab:preference}, which showed similar results, with the Original and Retrieval being considered equal, followed closely by Random, and Generative a distant fourth.

Qualitative analysis showed that the responses of the Retrieval are coherent despite the simple representation of dialogue context. 
Also, we found that, while large-scale pre-trained language models can be adequate in responding to general queries, they fail to produce good responses where more advanced vocabulary and reasoning are required.

Here we demonstrated methods for generating cohesive responses for group deliberation. That said, there is a reasonable question on whether these responses are also useful for the conversation, i.e. can they contribute towards improving the group's deliberation? One way to address that, in future dialogue system development, is to leverage DeliData's numerical measure of success to filter which utterances are contributing towards a constructive conversation, and which are not.

\section{Case study on 
collaborative debunking of deepfakes}
\label{sec:deepfake}

\begin{table}[]
    \centering
    \begin{tabular}{l|c|c}
    \textbf{Utterance} & \textbf{Type} & \textbf{Role} \\ \hline
    \multicolumn{1}{p{10cm}|}{And the punctuation here is weird too so maybe that is a clue to that potentially being something so do we do we agree on paragraph three, is our answer.}  & NPD & Reasoning \\ \hline
    Yes, all right um what kinds of reasons lacks common sense. & Probing & Reasoning \\ \hline
    I was not even sure what that means. Contradicts previous sentences. & NPD & Reasoning \\ \hline
    I agree with that statement. & NPD & Agree \\ \hline
    \end{tabular}
    \caption{Abridged annotated example from the dataset of collaborative deepfake debunking. NPD stands for Non-probing deliberation}
    \label{tab:deepfkate}
\end{table}

We have so far studied the Wason card selection task, which is well-studied and characterized by previous work on group decision-making. While we have argued the use of the Wason card selection task is appropriate due to its abstract nature that does not require prior knowledge, 
\newtext{we want to examine}
whether the patterns learned on Wason are transferable to other tasks, closer to real-life scenarios. To this end, we investigate how accurately our classifiers trained on \delidata (introduced in 
section~\ref{sec:delidata_prediction}) can perform in an out-of-domain task.

For this purpose, we use the dataset by Unchendu et al.~\cite{dongwon} that contains group discussions, where people collaborate to determine whether an image they are seeing is a deepfake or not. An example snippet is presented  \newtext{in Table} \ref{tab:deepfkate}. We \newtext{manually} annotated 110 utterances from \newtext{this} 
dataset with the annotation scheme proposed in this work. The label distribution of the annotated examples is presented \newtext{in Table}~\ref{table:deepfake_training}.

\begin{table}[t!]
\centering
    \begin{tabular}{|l||l|l|l||l|}
    \hline 
    \textbf{} & \textbf{Probing} & \multicolumn{1}{|p{2cm}|}{\centering \textbf{Non-probing deliberation}} & \textbf{None} & \textbf{Total} \\
\hline \hline 
    Moderation & 6 & 0 & & 6 \\ \hline 
    Reasoning & 1 & 46 & & 47 \\ \hline 
    Solution & 0 & 5 & & 5 \\ \hline 
    Agree & 0 & 33 & & 33 \\ \hline 
    Disagree & 0 & 2 & & 2 \\ \hline \hline 
    \textbf{Total} & 7 & 86 & 17 & 110 \\ \hline
\end{tabular}
\caption{Label counts in the top two levels of the deepfake annotated data}
\label{table:deepfake_training}
\end{table} 

We use this subset of the data to benchmark the accuracy of the classifiers introduced in section~\ref{sec:delidata_prediction}. \newtext{For these experiments, we selected the best-performing models, namely the neural hierarchical classifier.} On the first level of annotation (Probing vs Non-probing deliberation vs None) we achieve an accuracy of 71\% (compared to 87\% on DeliData). On the second level of annotation, the classifiers achieve a combined performance of 64\% accuracy (compared to 86\% on DeliData). 
This performance is encouraging, and while we note that it is lower than on the original DeliData set, 
we would like to highlight that the Wason card selection task and deepfake debunking are very different tasks, with very different vocabularies. 

In order to get a better understanding of the model, we performed an error analysis study. We note that the biggest drop in performance is for the \textbf{Solution} labels (0.17 F1 score vs 0.83 on DeliData). A solution in the case of the Wason card selection task is focused on which cards should be turned in order to verify a rule; on the other hand, for the deep fake data, participants are proposing whether an image is credible or not. As a result, solution utterances differ substantially in terms of the language used, and this performance drop is \newtext{expected}. 
The performance on all other labels \newtext{remains} 
high, 
e.g.\ for \textbf{Reasoning}, 
which had an F1 score of 0.75, compared to 0.91 on DeliData. This shows that even on \newtext{a} substantially different task, people use similar argumentation structure, 
\newtext{expressions}
and deliberation patterns to express their reasoning. An example snippet of an automatically annotated conversation is presented in appendix~\ref{app:deepfake}. 

Given these results, two conclusions are important for future work. Firstly, we show that the annotation scheme introduced in this paper applies to other tasks, and can benefit research investigating collaboration and deliberation. Secondly, even though the Wason task is an abstract 
task \newtext{devised} 
with the purpose to study decision-making, the patterns learned on the DeliData corpus are transferable to real-world collaborative scenarios such as deepfake debunking.

\section{Conclusions and Future Work}
\label{sec:future_work}

In this work, we introduced a dataset containing conversations where a group of participants collaborate in order to solve a task.
Furthermore, we proposed an annotation schema and annotated corpus that capture key elements of group deliberation, such as probing. 
In order to evaluate the dataset and the annotation scheme we performed 4 types of modelling experiments. First, we showed that we can build a classifier that predicts the annotated labels with high accuracy.  Following, we investigated methods for predicting conversational success based on dialogue features and annotations.  Then, we show that one of the modules for a future dialogue system can be addressed with a retrieval method. 
Finally, we showed that the classifier and the annotation scheme are transferring well to an out-of-domain dataset.
Given the resources and the conclusions from our experiments, we believe that this paper is a step towards addressing the call for ``discourse optimization'' of \citet{vecchi-etal-2021-towards}.


Two main research directions can be addressed \newtext{in future research}. 
First, while this paper performs an analysis on what contributes towards improved group decision-making (referred to as performance gain), future work should perform a more in-depth analysis of the conditions under which a group \textit{performs better than the sum of its parts}. For example, it can be processed with tools developed in argumentation mining \cite{Zhang2022EnhancingLA} and discourse parsing 
\cite{Liu2021ImprovingMD} in order to provide insights in how these relate to problem-solving deliberation. 
Secondly, this dataset can be analysed to test theories of the dynamics of group deliberation and develop 
 dialogue agents that could be used to improve group decision-making in numerous setups, for example debating groups, project meetings, etc. Such dialogue agents could be decomposed into three modules - determining intervention timing,
determining intervention type (i.e.\ moderation, probing for reasoning) and generating a probing utterance. In this work we introduced an adequate approach for coherent probing generation (Section~\ref{ssec:generation}), however determining the timing and the type of intervention is left for future work.

\section{Ethics Statement}
In this work, we present a corpus containing conversations, where participants collaborate to solve a cognitive task. Details on our setup and ethical considerations are presented in Section \ref{sec:experimental_setup} and appendices \ref{app:reproducibility_insructions} and \ref{app:mturk}, but in this section we will reiterate  the most important points.

We collected our dataset using the crowd-sourcing platform MechanicalTurk and in-lab volunteers for the initial experiments. Participants gave informed consent to their participation, and we told them the purpose of the study and that the transcripts of the dialogues would be collected and used for further research. The only language used in our dataset is English. Participants were free to withdraw at any time. We asked participants not to share any personal information, and as part of quality control, we have removed any instances of such (like the city they were living in, or the institution they were studying in). We asked the participants not to use any offensive language, and as part of the quality control, we verified whether this is the case, fortunately not finding any such instances. 
When recruiting participants, we selected adult participants from countries where English is a primary language and where MechanicalTurk operated at the time of collection: US, Canada, UK, Ireland, \newtext{and} Australia. Besides that, we did not put any restrictions on (nor have a record of) participants' exact age, gender, nationality, race, political leaning, education, etc.

Crowd workers were paid on average between £12/hour and £18/hour (approx. \$16.46/h-\$24.68/h), depending on their time of participation and whether they solved the task correctly. This is well above the UK's living wage (£8.91/hour), as well as the minimum wage in the US (\$7.25) \footnote{https://www.dol.gov/general/topic/wages/minimumwage}.
Moreover, in cases where we were unable to start the data collection (due to inactive users for example), we paid the participants for their time.

For our human evaluation experiments, we recruited participants from Prolific. We put similar qualification requirements as on MechanicalTurk, namely, minimum age of 18, fluent in English, and minimum approval rate of 90\%. We paid annotators in the same pay range as on MechanicalTurk, averaging £14.25/hr (19.5\$/h).

The full experimental design was approved by the ethics committee of the authors' institution. 
We will release the DeliData corpus under Creative Commons 4.0. 

\par

Future work may be needed to evaluate whether this dataset would apply to other types of problem-solving (for example in a business setting).

\section*{Acknowledgments}

The authors would like to acknowledge the support of the Isaac Newton Trust and Cambridge University Press in creating this dataset. Georgi Karadzhov is supported by EPSRC doctoral training scholarship. Tom Stafford and Andreas Vlachos are supported by the EPSRC grant no.\ EP/T023414/1: Opening Up Minds.

\bibliographystyle{ACM-Reference-Format}

\bibliography{anthology,custom, additional_citations}
\appendix

\clearpage

\section{Reproducibility Checklist}
\subsection{Data Collection - Participant Instructions}
\label{app:reproducibility_insructions}
Participants are given the following description of the task and experiment:

\begin{enumerate}
\item You will be part of a small-group chat (3-5 people), where you will try solving a puzzle.
\item Finish the task by yourself
\item Participate in a group discussion (via the chat), collaborate with the other participants and try to find the best solution together. Give your best effort both in solving the task and in the group discussion.
\item You are expected to participate actively in the conversation for at least 5 minutes.

\item Based on the discussion and arguments you had, submit the revised task solution again. You can submit the same answer if you believe it's the correct one.

\item \textbf{Task:} Each of the 4 cards below has a letter on one side and a number on the other. Which card(s) do you need to turn to test the rule:
All cards with vowels on one side have an even number on the other.
NB: Select ONLY the card(s) required to verify the rule. Most people get this task wrong.

\item Please remember that these transcripts may be used in future research, and therefore you have the right to withdraw from this study at any given time. To do so, press the ``Leave room'' button above. Please ensure you do not use any offensive language or disclose any personal information which would make you identifiable to others as it’s important that your anonymity is maintained. Any information which may reveal your identity will be deleted from this chat.
\end{enumerate}

\subsection{Data Collection: Mechanical Turk Modifications}
\label{app:mturk}
We recognise that collecting data on MechanicalTurk, we will face more challenging conditions compared to a controlled lab setup. Moreover, by design, MechanicalTurk is providing a platform for a single person to complete a task. As we aim at collecting group dialogues we applied to following recruitment protocol that enables synchronous data collection between multiple turkers: 

\begin{enumerate}
\item \textbf{Room Routing}.
Every crowd worker that joins our task is routed to a group that is recruiting participants or if none available - creates a new room. As we recognise, that some participants might leave after joining a room, we identified the following 3 room states: 
\begin{enumerate}
\item \textbf{Recruiting}: if the room has less than 3 active participants, a new participant can join at any time
\item \textbf{Final Call}: After there are at least 3 people in the room, a 1-minute timer starts, which allows for up to 2 more participants to join. By allowing more than 3 people to join, we mitigate the effect of inactive or leaving participants.
\item \textbf{Ready to Start}: Once the \textbf{final call} timer elapses, the game is ready to start.  
\end{enumerate}

\item \textbf{Crowd worker requirements}. 
To get high-quality data collection, the crowd workers participating in our task should meet the following conditions:
\begin{enumerate}
    \item Complete a simple reading comprehension test
    \item Fluency in English, which is established by being a resident of countries where English is an official language
    \item Have more than 95\% success rate on previous crowd-sourcing tasks
    \item Have completed at least 1000 tasks on Mechanical Turk
\end{enumerate}

\item \textbf{Notifications}.
Sometimes it takes a while for a group of 3 people to be ready, and, naturally, some of the participants may be inactive while waiting. To ensure that everyone is online, when the group is ready to start, there are audible notifications during key phases of the experiment, as well if someone is being inactive or not responsive during the game.   

\item \textbf{Quality Control}. We perform two kinds of quality control over the collected data. Initially, we automatically exclude all conversations that either have only a single participant in them or have less than 10 messages. Then, each conversation is manually checked, to ensure that no personal information was shared. Finally, we excluded conversations based on poor quality, i.e.\ when participants are not discussing the task at all. That said, participants are still getting paid if the conversation was excluded to no fault of their own.
\end{enumerate}

\subsection{Predicting Performance Gain}
\label{app:reproducibility_performance_gain}
To encourage reproducibility we will describe in details how we predict performance gain. 

\textbf{Conversation Statistics (9 features)}: Number of participants in the chat, total number of messages, average number of messages per player, average number of tokens per player, total unique tokens, average unique tokens per player, participants' individual performance, diversity in participants' individual solutions, and group consensus.

\textbf{Participation Dynamics (13 features)}. In the context of this work, we built a solution and participation tracker. Knowing the cards, presented to the participants, we track each solution proposal, as well as per participant change of solution. We do this by applying a simple rule-based system - if the message mentions one or more of the cards we save this as participant's solution proposal. Next time the same participant proposes a different solution we mark this event as a solution change.

Complimentary to the solution tracker, we also keep a record of how actively each participant engages in the discussion. We identify 4 categories of participation, based on how many messages each player issued - 0, 0-20, 40-50, 50-100 \%. Thus we are able to record both more silent users, and those who participate more than the rest of the group.

That said, we extract the following features: Number of solution changes (normalised by the number of messages), The 4 categories of participation at 20/50/all messages. 

\textbf{Annotation Features}
Here we considered 2 types of features related to the annotations introduced in this manuscript. First, we consider \textbf{Annotation Statistics} (20 features) which contain what proportion of each annotation label is present in the conversation. Secondly, we consider \textbf{Annotation N-gram features} (500) in order to capture some temporal information in the annotation patterns, we consider each conversation as sequence of annotations. Then, we use an n-gram vectoriser from \cite{scikit-learn}, with parameters - ngrams: [1-5], max features 500.

\textbf{StreetCrowd Features}
For more details, please refer to \cite{niculae2016conversational}.
\begin{itemize}
    \item \textbf{Interaction Features} (6 features). These features are calculated based on the whole conversation (rather than on an individual message). First, \cite{niculae2016conversational} include language matching on stopword, token and POS tag levels. Further, the interaction features capture agreement and disagreement markers in words. 
    
    \item \textbf{Linguistic Features} (15 features). These are message level features, that capture specific linguistic phenomena: message length (and it's variation), psycholinguistic features from LIWC \cite{Tausczik2010ThePM}, task specific jargon, and POS patterns.
\end{itemize}

\par
\textbf{Model Selection and Hyperparameter Search}.
Due to the relatively small size of the dataset, and the high information load of each conversation (large number of utterances), the selection of an appropriate model is a challenging endeavour. In our experiments, we found out that most models are either unable to generalise well or are very unstable in terms of performance. Models that performed poorly in either generalisation or stability were: Linear Regression, Support Vector Machine (both linear and RBF kernels), RandomForest, K-Nearest Neighbour, and a multilayer perceptron. Thus, we selected a decision tree, as it is a fairly stable model by design, and it allows us to analyse variability between different runs of the model. We performed hyperparameter search with the following parameters: Max Depth: [2, 3, 5, 7 (selected), 20, max] and Min Samples per leaf: [1, 2, 3, 5 (selected), 10]. Total number of parameter tuning runs - 30. The best model is selected based on model accuracy and stability. Due to the size of the model and the dataset, the hyperparameter search does not require any special infrastructure and the training time is negligible.

\subsection{Packages used}
For training and evaluation of the performance gain we used \cite{scikit-learn} version 1.0.2. For general language tasks and featurisers we used NLTK \cite{bird2009natural} version 3.5, Spacy \cite{spacy2} version 2.3.2. For generative experiments, we used DialoGPT-large from HuggingFace's transformers version 4.11.3.

For evaluation, we used BertScore \cite{bertscore} version 0.3.11, Sentence Transformers version 2.1.0.


\clearpage
\section{Example of a constructive and non-constructive conversation}
\label{app:annotated}
\begin{table}[h]
    \centering
    \scalebox{0.89}{
    \begin{tabular}{|l|l|c|c|l|} \hline
\textbf{User}  & \textbf{Utterance} & \textbf{Is Probing}  & \textbf{Role}  & \textbf{Additional Labels} \\ \hline 
Alpaca  & What did everybody put?  & Probing  & Moderation  & \\ \hline 
Leopard  & I put 6 and S, how about you?  & NPD  & Solution  & complete\_solution \\ \hline 
Alpaca  & \multicolumn{1}{|p{6cm}|}{ Oh, i thought we could only chose one card. I chose A } & NPD  & Solution  & complete\_solution \\ \hline 
Alpaca  & Why did you choose  & Probing  & Reasoning  &  \\ \hline 
Tiger  & \multicolumn{1}{|p{6cm}|}{ I put 6 - to see if has a vowel on the other side A to see if it has an even number and 7 to see if it has a consonant}  & NPD  & Reasoning  & complete\_solution \\ \hline 
Alpaca  & 6 and S  & NPD  & Solution  & complete\_solution \\ \hline 
Tiger  & I mean a vowel on 7  & NPD  & Reasoning  & partial\_solution \\ \hline 
Tiger  & \multicolumn{1}{|p{6cm}|}{ as if it is a vowel the rule wouldn't apply}  & NPD  & Reasoning  & partial\_solution \\ \hline 
Tiger  & \multicolumn{1}{|p{6cm}|}{ @Alpaca why do you think you need to turn s?}  & Probing  & Solution  & \multicolumn{1}{|p{3.5cm}|}{specific\_addressee, partial\_solution} \\ \hline 
Leopard  & \multicolumn{1}{|p{6cm}|}{ Okay I put 6 because I thpught we need to check if there's a vowel on the other side, and then S to make sure there's not an even number on that}  & NPD  & Reasoning  & complete\_solution \\ \hline 
Alpaca  & No i would only turn A  & NPD  & Disagree  & complete\_solution \\ \hline 
Alpaca  & \multicolumn{1}{|p{6cm}|}{ i would not choose 6 as the rule is not whether all even numbers have a vowel on the back, its if all vowels have an even number on the back}  & NPD  & Reasoning  & complete\_solution \\ \hline 
Leopard  & \multicolumn{1}{|p{6cm}|}{ Actually yeah I change my answer to A and 7}  & NPD  & Agree  & complete\_solution \\ \hline 
Tiger  & \multicolumn{1}{|p{6cm}|}{ Actually - do we need 6? it doesn't matter if it has a vowel or not}  & NPD  & Solution  & partial\_solution \\ \hline 
Alpaca  & so definitely A...  & NPD  & Solution  & partial\_solution \\ \hline 
Alpaca  & and i think 7  & NPD  & Solution  & partial\_solution \\ \hline 
Leopard  & \multicolumn{1}{|p{6cm}|}{ Don't we need to check 7 to make sure it doesn't have a vowel?}  & Probing  & Solution  & partial\_solution \\ \hline 
Alpaca  & Yes, I agree  & NPD  & Agree  &  \\ \hline 
Tiger  & Definettly A and I think 7 too  & NPD  & Solution  & complete\_solution \\ \hline 
Leopard  & Okay final answer A and 7 then?  & Probing  & Solution  & \multicolumn{1}{|p{3.5cm}|}{ solution\_summary, complete\_solution} \\ \hline 
Alpaca  & Do we all agree on 7 and A?  & Probing  & Solution  & \multicolumn{1}{|p{3.5cm}|}{solution\_summary, complete\_solution} \\ \hline 
Tiger  & yes  & NPD  & Agree  & \\ \hline 
    \end{tabular}
    }
    \caption{Constructive conversation ending in a correct solution}
    \label{table:example_annotation_positive}
\end{table}

\clearpage
\begin{table}[h]
    \centering
    \scalebox{0.89}{
    \begin{tabular}{|l|l|c|c|l|} \hline
\textbf{User}  & \textbf{Utterance} & \textbf{Is Probing}  & \textbf{Role}  & \textbf{Additional Labels} \\ \hline 
Beaver & I think we should check all four cards. & NPD & Solution & complete\_solution\\ \hline 
Bee & I am going with the last 2 & NPD & Solution & complete\_solution\\ \hline 
Narwhal & \multicolumn{1}{|p{6cm}|}{At the very least we should definitely include the 3rd card.} & NPD & Solution & partial\_solution\\ \hline 
Beaver & Ok, anything else? & Probing & Moderation & \\ \hline 
Bee & Why A? & Probing & Reasoning & \\ \hline 
Narwhal & \multicolumn{1}{|p{6cm}|}{The rule is that all cards with a vowel on one side have an even number on the other side.} & NPD & Solution & \\ \hline 
Narwhal & \multicolumn{1}{|p{6cm}|}{Well, our third card is a vowel to start with. We do not know what is on the other side of that card. If we flip our only apparent vowel and we find an even number, that is a pretty good indication to the rule right off the start.} & NPD & Reasoning & partial\_solution\\ \hline 
Beaver & ok & NPD & Agree & \\ \hline 
Bee & makes sense & NPD & Agree & \\ \hline 
Narwhal & \multicolumn{1}{|p{6cm}|}{None of the other cards would do us any good to flip them over because they are either an odd number or a consonant.}& NPD & Reasoning & \\ \hline 
Narwhal & So A is the way to go. & NPD & Solution & complete\_solution\\ \hline 
Bee & sounds good to me. & NPD & Agree & \\ \hline 
Beaver & A it is, & NPD & Agree & complete\_solution\\ \hline 
Bee & Thanks for the help, &  &  & \\ \hline 
Narwhal & Thanks for being willing to listen! &  &  & \\ \hline
    \end{tabular}
    }
    \caption{Non-constructive conversation}
    \label{table:example_annotation_negaitve}
\end{table}

\clearpage
\section{Examples of different approaches to generating utterances}
\label{app:example_NLG}

\begin{table}[h!]
    \centering
    \begin{tabular}{|c|l|}
        \hline
        \textbf{Context} & \\ \hline 
        Narwhal & Hello \\ \hline
        Dolphin & Hi \\ \hline
         \textbf{Original} & Anyone have any suggestion to a solution \\ \hline
         \textbf{Random} & Dolphin what did you select \\ \hline
         \textbf{Retrieval} & so what we are supposed to discuss about \\ \hline
         \textbf{Generative} & hey \\ \hline
    \end{tabular}
    \caption{Example of different methods for generating Probing-Moderation utterances}
    \label{tab:generative_example_moderation}
\end{table}

\begin{table}[h!]
    \centering
    \begin{tabular}{|c|l|}
        \hline
        \textbf{Context} & \multicolumn{1}{|p{12cm}|}{ but it says it might be as simple as we think and it seems pretty simple to put U and 2 as that is the vowel and the even number} \\ \hline
         \textbf{Original} & So is it 7 ? \\ \hline
         \textbf{Random} & so 2 , U , and 7\\ \hline
         \textbf{Retrieval} & So you think the 2 Card ? \\ \hline
         \textbf{Generative} & \multicolumn{1}{|p{12cm}|}{I concur} \\ \hline
    \end{tabular}
    \caption{Example of different methods for generating Probing-Solution utterances}
    \label{tab:generative_example_solution}
\end{table}


\clearpage

\section{Automatic annotation of Deepfake debunking transcripts}
\label{app:deepfake}
\begin{table}[h]
    \centering
    \scalebox{0.89}{
    \begin{tabular}{|l|l|c|c|l|} \hline
\textbf{Utterance} & \textbf{Is Probing}  & \textbf{Role}  \\ \hline 
 Why does it seem like it's generated from. & Probing & Reasoning \\ \hline 
I just said, it seems like the least related maybe. & NPD & Reasoning \\ \hline 
Potentially less related very succinct. & NPD & Reasoning \\ \hline 
P6 said to pick it. & NPD & Solution \\ \hline 
Third, one for this, because it seems unrelated. & NPD & Reasoning \\ \hline 
But it could be related. & NPD & Reasoning \\ \hline 
i'm not seeing a connection. & NPD & Reasoning \\ \hline 
What do you think Jackie. & Probing & Moderation \\ \hline 
Two yeah. & NPD & Solution \\ \hline 
Why do you think so. & Probing & Reasoning \\ \hline 
You know, it seems unrelated just like P5. & NPD & Reasoning \\ \hline 
spelling. & NPD & Solution \\ \hline 
Is a human generated. & NPD & Solution \\ \hline 
is not necessarily related to it could be, but it's well written it's well written clear and succinct number three. & NPD & Reasoning \\ \hline 
yeah one in three could be unrelated but we're not really yeah. & NPD & Reasoning \\ \hline 
    \end{tabular}
    }
    \caption{Automatically annotated example from the deepfake debunking dataset}
    \label{table:deepfake_autoannotation_example}
\end{table}

\end{document}